\documentclass[11pt,a4paper]{article}
\usepackage[hyperref]{emnlp-ijcnlp-2019}
\usepackage{times}
\usepackage{latexsym}

\usepackage{url}
\usepackage{natbib}
\usepackage{amsmath}
\usepackage{mathtools}
\usepackage{cleveref}
\usepackage{graphicx}
\usepackage{xcolor}
\usepackage{multirow}
\usepackage[normalem]{ulem}

\newcommand{\aline}[1]{#1}

\def\Plus{\texttt{+}}
\def\Minus{\texttt{-}}

\aclfinalcopy

\title{Why So Down? The Role of Negative (and Positive) Pointwise \\Mutual Information
in Distributional Semantics}

\author{Alexandre Salle$^{1}$ \quad Aline Villavicencio$^{1,2}$ \\
  $^{1}$Institute of Informatics, Federal University of Rio Grande do Sul (Brazil) \\
  $^{2}$School of Computer Science and Electronic Engineering, University of Essex (UK)\\
  {\tt alex@alexsalle.com \quad avillavicencio@inf.ufrgs.br }}

\date{}

\begin{document}
\maketitle
\begin{abstract}
  In distributional semantics, the pointwise mutual information ($\mathit{PMI}$) weighting of the cooccurrence matrix performs far better than raw counts. There is, however, an issue with unobserved pair cooccurrences as $\mathit{PMI}$ goes to negative infinity. This problem is aggravated by unreliable statistics from finite corpora which lead to a large number of such pairs. A common practice is to clip negative $\mathit{PMI}$ ($\mathit{\Minus PMI}$) at $0$, also known as Positive $\mathit{PMI}$ ($\mathit{PPMI}$). In this paper, we investigate alternative ways of dealing with $\mathit{\Minus PMI}$ and, more importantly, study the role that negative information plays in the performance of a low-rank, weighted factorization of different $\mathit{PMI}$ matrices.
  Using various semantic and syntactic tasks as probes into models which use either negative or positive $\mathit{PMI}$ (or both), we find that most of the encoded semantics and syntax come from positive $\mathit{PMI}$, in contrast to $\mathit{\Minus PMI}$ which contributes almost exclusively syntactic information. Our findings deepen our understanding of distributional semantics, while also introducing novel $PMI$ variants and grounding the popular $PPMI$ measure.
\end{abstract}

\section{Introduction}
Dense word vectors (or embeddings) are a key component in modern NLP architectures for tasks such as sentiment analysis, parsing, and machine translation. These vectors can be learned by exploiting the distributional hypothesis \cite{Harris1954}, paraphrased by \citet{firth1957synopsis} as \emph{``a word is characterized by the company that it keeps''}, usually by constructing a cooccurrence matrix over a training corpus, re-weighting it using Pointwise Mutual Information ($\mathit{PMI}$) \cite{church1990word}, and performing a low-rank factorization to obtain dense vectors. 

Unfortunately, $\mathit{PMI}(w,c)$ goes to negative infinity when the word-context pair $(w,c)$ does not appear in the training corpus. Due to unreliable statistics, this happens very frequently in finite corpora. Many models work around this issue by clipping negative $\mathit{PMI}$ values at $0$, a measure known as Positive $\mathit{PMI}$ ($\mathit{PPMI}$), which works very well in practice. An unanswered question is: \emph{``What is lost/gained by collapsing the negative $\mathit{PMI}$ spectrum to $0$?''}.
\aline{Understanding which type of information is captured by $\mathit{\Minus PMI}$ can help in tailoring models for optimal performance}.

In this work, we attempt to answer this question by studying the kind of information contained in the negative and positive spectrums of $\mathit{PMI}$ ($\mathit{\Minus PMI}$ and $\mathit{\Plus PMI}$). We evaluate weighted factorization of different matrices which use either $\mathit{\Minus PMI}$, $\mathit{\Plus PMI}$, or both on various semantic and syntactic tasks. Results show that $\mathit{\Plus PMI}$ alone performs quite well on most tasks, capturing both semantics and syntax, in contrast to $\mathit{\Minus PMI}$, which performs poorly on nearly all tasks, except those that test for syntax. Our main contribution is deepening our understanding of distributional semantics by extending \citet{firth1957synopsis}'s paraphrase of the distributional hypothesis to \emph{``a word is not only characterized by the company that it keeps, but also by the company it rejects''}. Our secondary contributions are the proposal of two $PMI$ variants that account for the spectrum of $\mathit{\Minus PMI}$, and the justification of the popular $PPMI$ measure.

In this paper, we first look at related work ($\S$\ref{sec:related}), then study $\mathit{\Minus PMI}$ and ways of accounting for it ($\S$\ref{sec:model}), describe experiments ($\S$\ref{sec:materials}), analyze results ($\S$\ref{sec:results}), and close with ideas for future work ($\S$\ref{sec:conclusion}).

\section{Related Work}
\label{sec:related}
There is a long history of studying weightings (also known as association measures) of general (not only word-context) cooccurrence matrices; see \citet{manning1999foundations,jurafsky2000speech} for an overview \aline{and \citet{Curran:2002} for comparison of different weightings}. 
\citet{Bullinaria2007}  show that word vectors derived from $\mathit{PPMI}$ matrices perform better than alternative weightings for word-context cooccurrence. In the field of collocation extraction, \citet{bouma2009normalized} address the negative infinity issue with $\mathit{PMI}$ by introducing the normalized $\mathit{PMI}$ metric. 
\citet{Levy2014a} show theoretically that the popular Skip-gram model \citep{Mikolov2013} performs implicit factorization of shifted $\mathit{PMI}$.

Recently, work in explicit low-rank matrix factorization of $\mathit{PMI}$ variants has achieved state of the art results in word embedding. GloVe \citep{Pennington2014} performs weighted factorization of the log cooccurrence matrix with added bias terms, but does not account for zero cells. \citet{shazeer2016swivel} point out that GloVe's bias terms correlate strongly with unigram log counts, suggesting that GloVe is factorizing a variant of $\mathit{PMI}$. Their SwiVel model modifies the GloVe objective to use Laplace smoothing and hinge loss for zero counts of the cooccurrence matrix, directly factorizing the $\mathit{PMI}$ matrix, sidestepping the negative infinity issue. An alternative is to use $\mathit{PPMI}$ and variants as in \citet{W14-1503, E14-1025,P16-3009,Salle2016,xin2018batch}. \aline{However, it is not clear what } 
is lost by clipping the negative spectrum of $\mathit{PMI}$, which makes the use of $\mathit{PPMI}$, though it works well in practice, seem unprincipled.

In the study of language acquisition, \citet{regier2004learning} argue that indirect negative evidence might play an important role in human acquisition of grammar, but do not link this idea to distributional semantics.

\section{PMI \& Matrix Factorization}
\label{sec:model}
\textbf{PMI:} A cooccurrence matrix $M$ is constructed by sliding a symmetric window over the subsampled \cite{Mikolov2013} training corpus and for each center word $w$ and context word $c$ within the window, incrementing $M_{wc}$. $\mathit{PMI}$ is then equal to:
\begin{align}
  \mathit{PMI}(w,c) = \log \frac{M_{wc} \; M_{**}}{ M_{w*} \; M_{*c} }
\end{align}
where * denotes summation over the corresponding index. To deal with negative values, we propose clipped $\mathit{PMI}$, 
\begin{align}
 \mathit{CPMI}_z (w, c) = max(z, PMI(w,c))
\end{align}
which is equivalent to $\mathit{PPMI}$ when $z = 0$.

\textbf{Matrix factorization:} LexVec \cite{Salle2016} performs the factorization $M' = WC^\top$, where $M'$ is any transformation of $M$ (such as $\mathit{PPMI}$), and $W, C$ are the word and context embeddings respectively. By sliding a symmetric window over the training corpus (window sampling), LexVec performs one Stochastic Gradient Descent (SGD) step every time a $(w,c)$ pair is observed, minimizing 
\begin{align*}
  L_{wc} &= \frac{1}{2} (W_w C_c^\top - M'_{wc})^2 
\end{align*}
Additionally, for every center word $w$, $k$ negative words \cite{Mikolov2013} are drawn from the unigram context distribution $P_n$ (negative sampling) and SGD steps taken to minimize:
\begin{align*}
  L_{w} &= \frac{1}{2} \sum\limits_{i=1}^k{\mathbf{E}_{c_i \sim P_n(c)} (W_w C_{c_i}^\top - M'_{wc_i})^2 }
\end{align*}

Thus the loss function prioritizes the correct approximation of frequently cooccurring pairs and of pairs where either word occurs with high frequency; these are pairs for which we have more reliable statistics. 

In our experiments, we use LexVec
over Singular Value Decomposition (SVD) because a) Empirical results shows it outperforms SVD \cite{Salle2016}. b) The weighting of reconstruction errors by statistical confidence is particularly important for $\mathit{\Minus PMI}$, where negative cooccurrence between a pair of frequent words is more significant and should be better approximated than that between a pair of rare words. GloVe's matrix factorization is even more unsuitable for our experiments as its loss weighting --- a monotonically increasing function of $M_{wc}$ --- ignores reconstruction errors of non-cooccurring pairs.

\begin{figure}
  \includegraphics[scale=0.42]{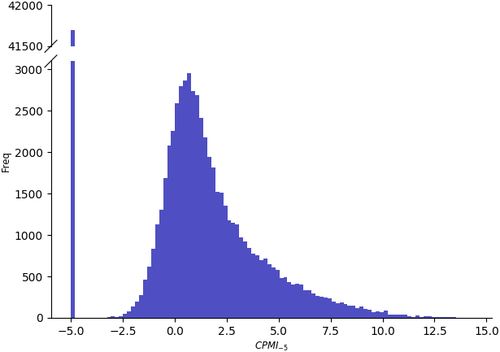}
  \caption{$\mathit{CPMI}_{\Minus 5}$ histogram (bucket width equal to $.2$) of $10^5$ sampled pairs using window sampling and negative sampling. Number of samples in interval: $[-5, -5] = 41695$, $(-5,0] = 11001$, $[-2, 0]=10759$, $(0, \infty)=47304$}\label{fig:hist}
\end{figure}

\textbf{Spectrum of PMI:} To better understand the distribution of $\mathit{CPMI}$ values, we plot a histogram of $10^5$ pairs randomly sampled by window sampling and negative sampling in \cref{fig:hist}, setting $z=-5$. We can clearly see the spectrum of $\mathit{\Minus PMI}$ that is collapsed when we use $\mathit{PPMI}$ ($z=0$). In practice we find that $z=-2$ captures most of the negative spectrum and consistently gives better results than smaller values so we use this value for the rest of this paper. We suspect this is due to the large number of non-cooccurring pairs ($41.7\%$ in this sample) which end up dominating the loss function when $z$ is too small. 

\textbf{Normalization:} We also experiment with normalized $\mathit{PMI}$ ($\mathit{NPMI}$) \cite{bouma2009normalized}:
\begin{align*}
  \mathit{NPMI}(w,c) = \mathit{PMI}(w,c) / -log(M_{wc}/M_{**})
\end{align*}
such that $NPMI(w,c) = -1$ when $(w,c)$ never cooccur, $NPMI(w,c) = 0$ when they are independent, and $NPMI(w,c) = 1$ when they always cooccur together. This effectively captures the entire negative spectrum, but has the downside of normalization which discards scale information. In practice we find this works poorly if done symmetrically, so we introduce a variant called $\mathit{NNEGPMI}$ which only normalizes $\mathit{\Minus PMI}$:

\begin{equation*}
  \begin{split}
  \mathit{NNEGPMI}(w,c) = \\
  \begin{cases*}
    \mathit{NPMI}(w,c) & if $\mathit{PMI}(w,c)<0$ \\
    \mathit{PMI}(w,c)        & otherwise
  \end{cases*}
  \end{split}
\end{equation*}

We also experimented with Laplace smoothing as in \citet{turney2003measuring} for various pseudocounts but found it to work consistently worse than both $\mathit{CPMI_z}$ and $\mathit{NNEGPMI}$ so we omit further discussion in this paper.

\section{Materials}
\label{sec:materials}

In order to identify the role that $\mathit{\Minus PMI}$ and $\mathit{\Plus PMI}$ play in distributional semantics, we train LexVec models that skip SGD steps when target cell values are $>0$ or $\leq 0$, respectively. For example, $-\mathit{CPMI}_{\Minus 2}$ skips steps when $\mathit{CPMI}_{\Minus 2}(w,c) > 0$. Similarly, the $\mathit{\Plus PPMI}$ model skips SGD steps when $\mathit{PPMI}(w,c) \leq 0$. We compare these to models that include both negative and positive information to see how the two interact.

We use the default LexVec configuration for all $\mathit{PMI}$ variants: fixed window of size $2$, embedding dimension of $300$, $5$ negative samples, positional contexts\footnote{Positional contexts account for the position of a context word relative to the target word--- e.g., $M_{wc_{\Minus 1}}$ is the number of occurrences of $c$ immediately to the left of $w$.}, context distribution smoothing of $.75$, learning rate of $.025$, no subword information, and negative distribution power of $.75$. We train on a lowercased, alphanumerical 2015 Wikipedia dump with $3.8$B tokens, discarding tokens with frequency $< 100$, for a vocabulary size of $303,517$ words.

For comparison, we include results for a randomly initialized, non-trained embedding to establish task baselines.

\textbf{Semantics:}
To evaluate word-level semantics, we use the SimLex \cite{hill2015simlex} and Rare Word (RW) \cite{Luong2013} word similarity datasets, and the Google Semantic (GSem) analogies \cite{Mikolov2013}. We evaluate sentence-level semantics using averaged bag of vectors (BoV) representations on the Semantic Textual Similarity (STSB) task \cite{cer2017semeval} and Word Content\footnote{By construction most probe words are content words, thus recovery relies on semantic information.} (WC) probing task (identify from a list of words which is contained in the sentence representation) from SentEval \cite{conneau2018you}. 

\textbf{Syntax:}
Similarly, we use the Google Syntactic analogies\footnote{Google Syntactic analogies are in fact morphological but many categories test for POS relations and are therefore syntactic in nature.} (GSyn) \cite{Mikolov2013} to evaluate word-level syntactic information, and Depth (Dep) and Top Constituent (TopC) (of the input sentence's constituent parse tree) probing tasks from SentEval \cite{conneau2018you} for sentence-level syntax. Classifiers for all SentEval probing tasks are multilayer perceptrons with a single hidden layer of 100 units and dropout of $.1$. Our final syntactic task is part-of-speech (POS) tagging using the same BiLSTM-CRF\footnote{Using https://github.com/zalandoresearch/flair} setup as \citet{Huang2015BidirectionalLM} but using only word embeddings (no hand-engineered features) as input, trained on the WSJ section of the Penn Treebank \cite{Marcus:1993:BLA:972470.972475}.

\section{Results}
\label{sec:results}

\begin{table*}
  \centering

\begin{tabular}{|c|c|c|c|c|c||c|c|c|c|}
  \hline 
  \multirow{2}{*}{\textbf{Model}} & \multicolumn{5}{c||}{\textbf{Semantic}} & \multicolumn{4}{c|}{\textbf{Syntactic}}\tabularnewline
  \cline{2-10} \cline{3-10} \cline{4-10} \cline{5-10} \cline{6-10} \cline{7-10} \cline{8-10} \cline{9-10} \cline{10-10} 
   & \textbf{SimLex} & \textbf{RW} & \textbf{GSem} & \textbf{STSB} & \textbf{WC} & \textbf{GSyn} & \textbf{POS} & \textbf{Dep} & \textbf{TopC}\tabularnewline
  \hline 
  \hline 
  $\mathit{\Plus PPMI}$ & \textbf{.377} & .352 & 56.1 & \uline{.622} & \textbf{74.1} & 50.3 & 92.2 & 30.5 & \uline{34.6}\tabularnewline
  \hline 
  $\mathit{\Minus CPMI_{\Minus 2}}$ & .164 & .231 & 3.6 & .402 & 22.7 & 7.1 & 89.6 & \textbf{32.7} & \textbf{34.7}\tabularnewline
  \hline 
  $\mathit{\Minus NNEGPMI}$ & .142 & .232 & 3.3 & .366 & 16.6 & 6.3 & 88.8 & \uline{32.4} & 34.1\tabularnewline
  \hline 
  \hline 
  $\mathit{PPMI}$ & \uline{.363} & \textbf{.459} & \uline{80.3} & .618 & 69.6 & 62.2 & \uline{92.5} & 29.0 & 30.5\tabularnewline
  \hline 
  $\mathit{CPMI_{\Minus 2}}$ & .355 & .432 & \uline{80.3} & .621 & 69.9 & \textbf{65.1} & \textbf{92.6} & 28.5 & 31.1\tabularnewline
  \hline 
  $\mathit{NPMI}$ & .322 & .437 & 63.5 & .578 & 58.0 & 58.0 & 92.1 & 29.2 & 31.4\tabularnewline
  \hline 
  $\mathit{NNEGPMI}$ & .360 & \uline{.439} & \textbf{80.7} & \textbf{.629} & 70.0 & \uline{64.2} & \textbf{92.6} & 27.2 & 30.3\tabularnewline
  \hline 
  \hline 
  Random & -.018 & -.026 & 0.0 & .453 & 0.3 & 0.0 & 55.2 & 17.9 & 5.0\tabularnewline
  \hline 
  \end{tabular}

  \caption{SimLex and RW word similarity: Spearman rank correlation. STSB: Pearson correlation. GSem/GSyn word analogy, POS tagging and WC, Dep, TopC probing tasks: \% accuracy. Best result for each column in bold, second best underlined.}
  \label{tab:senteval}
\end{table*}

All results are shown in \cref{tab:senteval}.

\textbf{Negative PMI:} We observe that using only $\mathit{\Minus PMI}$ (rows $\mathit{\Minus CPMI_{\Minus 2}}$ and $\mathit{\Minus NNEGPMI}$) performs similarly to all other models in POS tagging and both syntactic probing tasks, but very poorly on all semantic tasks, strongly supporting our main claim that $\mathit{\Minus PMI}$ mostly encodes syntactic information. 

Our hypothesis for this is that the grammar that generates language implicitly creates negative cooccurrence and so $\mathit{\Minus PMI}$ encodes this syntactic information. Interestingly, this idea creates a bridge between distributional semantics and the argument by  \citet{regier2004learning} that indirect negative evidence might play an important role in human language acquisition of grammar.  %

\textbf{Positive PMI:} The $\mathit{\Plus PPMI}$ model performs as well or better as the full spectrum models on nearly all tasks, clearly indicating that $\mathit{\Plus PMI}$ encodes both semantic and syntactic information. 

\textbf{Why incorporate -PMI?} $\mathit{\Plus PPMI}$ only falters on the RW and analogy tasks, and we hypothesize this is where $\mathit{\Minus PMI}$ is useful: in the absence of positive information, negative information can be used to improve rare word representations and word analogies. Analogies are solved using nearest neighbor lookups in the vector space, and so accounting for negative cooccurrence effectively repels words with which no positive cooccurrence was observed. In future work, we will explore incorporating $\mathit{\Minus PMI}$ only for rare words (where it is most needed).

\textbf{Full spectrum models:} The $\mathit{PPMI}$, $\mathit{CPMI_{\Minus 2}}$, and $\mathit{NNEGPMI}$ models perform similarly, whereas the $\mathit{NPMI}$ model is significantly worst on nearly all semantic tasks. We thus conclude that accounting for scale in the positive spectrum is more important than in the negative spectrum. We hypothesize this is because scale helps to uniquely identify words, which is critical for semantics (results on $WC$ task correlate strongly with performance on semantic tasks), but in syntax, words with the same function should be indistinguishable. Since $\mathit{\Plus PMI}$ encodes both semantics and syntax, scale must be preserved, whereas $\mathit{\Minus PMI}$ encodes mostly syntax, and so scale information can be discarded.

\textbf{Collapsing the negative spectrum:} The $\mathit{PPMI}$ model, which collapses the negative spectrum to zero, performs almost identically to the $\mathit{CPMI_{\Minus 2}}$ and $\mathit{NNEGPMI}$ models that account for the range of negative values. This is justified by 1) Our discussion which shows that $\mathit{\Plus PMI}$ is far more informative than $\mathit{\Minus PMI}$ and 2) Looking at \cref{fig:hist}, we see that collapsed values --- interval $(-5,0]$ --- account for only $11\%$ of samples compared to $41.7\%$ for non-collapsed negative values.

\section{Conclusions and Future Work}
\label{sec:conclusion}
In this paper, we evaluated existing and novel ways of incorporating $\mathit{\Minus PMI}$ into word embedding models based on explicit weighted matrix factorization\footnote{Code available at https://github.com/alexandres/lexvec}, and, more importantly, studied the role that $\mathit{\Minus PMI}$ and $\mathit{\Plus PMI}$ each play in distributional semantics, finding that \emph{``a word is not only characterized by the company that it keeps, but also by the company it rejects''}. 
In future work, we wish to further study the link between our work and language acquisition, and explore the fact the $\mathit{\Minus PMI}$ is almost purely syntactic to (possibly) subtract syntax from the full spectrum models, studying the frontier (if there is one) between semantics and syntax.

\section*{Acknowledgments}
This research was partly supported by CAPES and CNPq (projects 312114/2015-0, 423843/2016-8, and 140402/2018-7).

\bibliography{biblio}
\bibliographystyle{acl_natbib.bst}

\end{document}